# Reversible Image Merging for Low-level Machine Vision

**M. Kharinov**
St. Petersburg Institute for Informatics and Automation of RAS,
14_liniya Vasil'evskogo ostrova 39, St. Petersburg, 199178 Russia,
khar@iias.spb.su, www.spiiras.nw.ru

***Abstract:*** *In this paper a hierarchical model for pixel clustering and image segmentation is developed. In the model an image is hierarchically structured. The original image is treated as a set of nested images, which are capable to reversibly merge with each other. An object is defined as a structural element of an image, so that, an image is regarded as a maximal object. The simulating of none-hierarchical optimal pixel clustering by hierarchical clustering is studied. To generate a hierarchy of optimized piecewise constant image approximations, estimated by the standard deviation of approximation from the image, the conversion of any hierarchy of approximations into the hierarchy described in relation to the number of intensity levels by convex sequence of total squared errors is proposed.*

## 1. INTRODUCTION

Nowadays a new problem of automating the creation of artificial intelligence applications arises. This problem is solved, at least in the project PPAML (Probabilistic Programming for Advancing machine learning) of the US agency DARPA (Defense Advanced Research Projects Agency), which is scheduled for 2013-2017[1]. In the field of machine vision it seems necessary to create a unified software tool for the detection of an object hierarchy. Such tool should help the inexperienced programmer to select the «objects of interest» among many objects found in the image by computer.

Stumbling block for computer object detection is to define the object regardless of the image content and without preliminary machine learning. To avoid this difficulty it seems possible to treat the objects as the clusters of pixels of *optimal* approximations, which minimally differ from the image in the standard deviation $\sigma$ or total squared error $E = 3N\sigma^2$, where the coefficient 3 is the number of color components in the image. However, although such definition is based on the classic cluster analysis [1], the opportunities to minimize of the total squared error $E$ (the *approximating error*, for conciseness) in image processing domain are far from being exhausted, especially, in the task of multiple optimization for each number of pixel clusters.

So, continuing [2], we aim to really minimize the approximation errors in the generalized task of pixel clustering and image segmentation, as well as to create and implement the unified software for automatic object detection in the primary stage of image recognition and others processing tasks.

## 2. PROBLEM STATEMENT

In being created model for hierarchical pixel clustering and image segmentation the obtaining of hierarchical sequence of optimized approximations of $N$ pixels partitioned into each number of clusters from 1 to $N$ is treated.

The sequence of optimal approximations of $g$ clusters is described by the monotonous sequence of approximating errors $E_g$ that non-strictly decrease with the growth of the number of clusters $g$ from 1 to $N$ from the maximum value in case of sole cluster to zero when all pixels assigned to different clusters. The characteristic property of $E_g$ sequence is convexity:

$$E_g = \frac{E_{g-1} + E_{g+1}}{2}, \quad g = 2, 3, ..., N-1, \qquad (1)$$

causing the non-strict growth of increment $\Delta E$, along with $g$ decrease and increase of approximating error $E$.

The sequence of optimized approximations should accurately simulate a sequence of optimal approximations. Therefore, for a close similarity the convexity of target sequence of approximations is also required.

Hierarchical approximations that correspond to a convex sequence of $E_g$ values depending on the number of clusters $g$ are called *quasioptimal* [3,4].

Since the sequence of optimal approximations in general is non-hierarchical, then the sequence of quasioptimal approximations is not uniquely determined as in Fig.1.

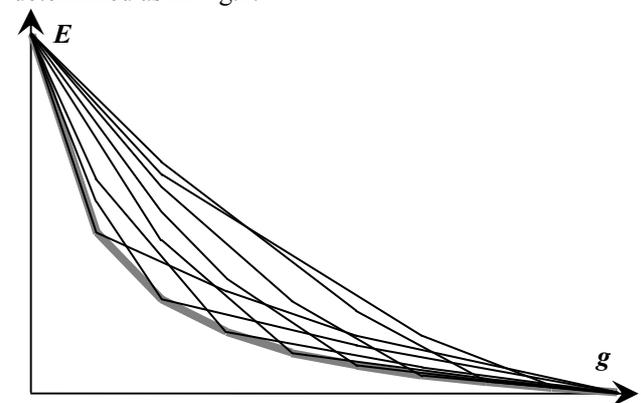

**Fig.1 – Simulating of optimal approximations**

Fig. 1 graphically illustrates the approximating error depending on the number of clusters in the image

---

approximations. Lowest bold gray curve corresponds to the optimal image approximations. Solid curves correspond to the target quasioptimal approximations that match the optimal at least for three cluster numbers. All the curves diverge from a common point for $g=1$, corresponding to the image approximation of a single cluster, and for $g=N$ converge at one point describing image approximation of $N$ clusters, each containing a single pixel. All curves shown in Fig. 1 are convex. Based on experiments with examples of gray images [2][2] it may be supposed that target quasioptimal approximations for color images are described by the curves passing through a series of optimal values and intricately intertwined depending on the image content. *Nearly optimal* curves within the deviations of quasioptimal curves of Fig. 1 from the optimal values lie in a quite narrow band over the optimal curve. So, in order to reliably appreciate and implement the effect of pixel clustering optimization, first of all it is highly desirable to increase the effectiveness of calculating of optimized approximations for color images.

To do this would be useful to pose and solve the task of converting of any hierarchical sequence of approximations into a sequence of quasioptimal approximations, as well as to minimize the approximation error for a fixed number of pixel clusters as in Fig. 1. The latter problem is standard, and it's usually declared in the practical applications of *K*-means method as well as its numerous modifications [1,5,6]. The first problem looks rather theoretical than practical. But it is the task of the conversion of any pixel clustering or image segmentation into quasioptimal one turns out most fruitful to unify the software.

### 3. IMAGE AND OBJECT NOTIONS

In the developed model of quasioptimal image approximations [3,4] a computer processing of the original image is represented as follows.

Let's consider that an image should be hierarchically structured. Let's agree that the original image initially is divided into pixels, which are treated as elementary images and are capable to reversibly merge with each other. Then during the processing the original image is divided into a number of structured sub-images that finally merge into a single completely structured image.

As a result of discussed processing, a dichotomous set of $2N-1$ pixel clusters, in special case, image segments, and a sequence of $N$ image approximations with each number of clusters or segments from 1 to $N$ is generated and stored in RAM in terms of Sleator-Tarjan dynamic trees without excessive memory usage [3,7,8].

The purpose of the model [3,4] is only producing the resultant hierarchical image approximations and pixel clusters or image segments for post-processing.

---

[2] See also http://oogis.ru/index.php/en/technologies/56-the-optimal-segmentation-dataset

Since any cluster of pixels may be treated as an image, the model does not involve any restrictions, but simply ascribes to an image a dichotomous cluster structure.

*Object*[3] in the model is considered to be an element of the image, i.e. one or another pixel clusters, found in the image by the computer. The image is a maximal object that includes other objects detected inside of given area of the original image. Completely structured original image includes all 2N-1 available objects detectable by computer. Intermediary structured original image, divided into several nested sub-images, contains only these sub-images and the objects detected in each of them. Unlike the image the notion of an object does not imply that any cluster of pixels may be treated as an object, as the dichotomous object hierarchy is calculated for the given image by the computer, in compliance with the logic of the problem statement.

### 4. REVERSIBLE IMAGE MERGING

Let $I_1$ and $I_2$ be the average intensities for the sub-images $S_1$ and $S_2$, respectively. Let $n_1$, $n_2$ be the corresponding numbers of pixels. Then the increment $\Delta E_{\text{merge}} \equiv E(S_1 \cup S_2) - E(S_1) - E(S_2)$ of the approximating error $E$ caused by the merging of specified sub-images along with reduction of their number per unit is given by the formula:

$$\Delta E_{merge} = \frac{n_1 n_2}{n_1 + n_2} \|I_1 - I_2\|^2 , \qquad (2)$$

where $\|I_1 - I_2\|^2$ is the square of the Euclidean distance between the three-component color values of averaged intensity.

It is the expression (2) used to iteratively calculate the hierarchical sequence of quasioptimal image approximations by Ward's method [9], that are produced according to the criterion:

$$\Delta E_{merge} = \min , \qquad (3)$$

whereby at each step of the hierarchy generation, the merging of clusters causing the minimal increment of the approximation error is performed.

The same estimation and criterion (2-3) are preferably used to iterative segmenting of an image by pair wise merging of only adjacent segments, as in Mumford Shah segmentation model [3,4,10-12].

Since the model of quasioptimal approximations allows any sub-images, then it imposes no a priori constraints to merging of the images. Therefore, image merging, in principle, can be carried out in any order, or in accordance with any model of pixel clustering or image segmentation. Moreover, in the framework of reversible computing technique [13], the reversibility of image merging operation is supported. This means that if the images $S_1$ and $S_2$ merge into the image $S_3$, then the latter becomes available for reverse dichotomous dividing operation just into the images $S_1$ and $S_2$, causing the negative or zero increment of approximating error:

$$\Delta E_{divide}(S_3) = -\Delta E_{merge}(S_1, S_2). \qquad (4)$$

---

[3] The term «*object*» is used as a synonym for the term «sub-image» and treated mainly in non-modifiable state, in particular to denote the resultant pixel clusters.

When merging, images transform into structural elements referred to as objects. Conversely, the dichotomous division of an image in twain causes its disintegration into objects. The convexity property of approximating errors for nested objects is expressed by the non-strictly decreasing of non-negative drop $|\Delta E_{divide}|$ of approximating error $E$ along with decreasing of nested objects

$$S_1 \subset S_2 \Rightarrow |\Delta E_{divide}(S_1)| \geq |\Delta E_{divide}(S_2)|, \qquad (5)$$

where the notation $S_2$ should be understood as the image to be produced.

From the standpoint of interpretation, the convexity claims (1), (5) simply means revealing of objects in descending order of contrast in different or in the same image locations. Convexity property is obviously not affected by the image dividing operation, but to preserve the convexity for merging operation the restructuring of composite image is required. Therefore, the image merging is carried out as a combined operation consisting of a uniting of the input hierarchies, followed by modification of obtained hierarchical sequence of sub-images to match the convexity conditions (1), (5).

To merge a pair of structured images into a single image, which is to be also structured, most likely, one can simply apply the Ward's method to a joint set of pixels. But to account that each of the two input sub-images is relied preliminary ordered, a more careful algorithm is used. This algorithm aims to keep the established order and does not change joint hierarchical sequence of sub-images, if they are described by the convex sequence of approximating errors. Otherwise, the computer detects wrong embeddings causing convexity violations, eliminates them by dividing a joint image into sub-images and then merges these sub-images into a single image by Ward's method. Since merging of the structured sub-images, in general, initiates novel convexity disorders, the detection of improper embeddings is performed again, partitioning into sub-images is crushed and the processing is iteratively repeated until obtaining the perfectly structured joint image.

Thus, if the available software supports reversible merging, storing, retrieving, analyzing and transforming of the hierarchy of ordinary clusters [3], then the transition to the advanced reversible merging of images, arranged in descending order of increment of the absolute values of the approximation errors along increasing of the number of sub-images, is provided by simple utilization of the combined operation of image merging and restructuring instead of cluster merging operation.

## 5. PIXEL CLUSTERING IMPROVEMENT

Our experience of segmentation regardless of predetermined image content, leads to the conclusion that conventional Ward's clustering [9] is not sufficiently used in the initial processing and streamlining of video data due to excessive computational complexity.

At first glance, mentioned challenge is overcome in the task of complete dichotomous structuring of original image by merging of only adjacent sub-images with few pixels in the first stage of image segmentation and subsequent clustering of the rest several sub-images with many pixels in the second, final stage of generating the hierarchically structured image that simulated by the sequence of hierarchical approximations and described by the convex sequence of approximating errors. In this case the resultant hierarchy of approximations will be described by piecewise convex curve. But upon closer experimental study it turned out that the correction of approximations to smooth the overall curve still requires heavy computation. To withdraw this problem of data adjustment by refinement of rough structured images obtained in the first stage of processing and at the same time to effectively minimize the approximating error for given cluster number the method presented in this section is developed.

This method of pixel clustering, in particular image segmentation for a given intermediate number of sub-images ensures that the maximal drop of the approximating error $\max |\Delta E_{divide}|$, caused by division in twain of certain one from all sub-images, would not exceed its minimal increment $\min \Delta E_{merge}$, caused by merging of certain sub-image pair chosen from all pairs of sub-images:

$$\min \Delta E_{merge} \geq \max |\Delta E_{divide}|. \qquad (6)$$

If the condition (6) is violated, then the division of the sub-image, which induces maximum drop of approximating error, is carried out. In the following step, the calculation of, in general case, a new pair of sub-images providing while merging the minimal increase of approximating error, is done. Next, a pair of found sub-images merges into one image by means of the combined operation of merging with the immediate restructuring of the joint image. Finally, the minimization process is either terminates, if criterion (6) is fulfilled, or resumed while (6) is not fulfilled.

Note, that merging of one of two parts of the sub-image divided in twain, to the other sub-image is envisaged in the above method.

Since in precedent versions [3,4,14] the discussed method was referred to as SI-method (abbreviated Segmentation Improving), then in current version we call it ASI-method (Advanced Segmentation Improving) to emphasis the utilization of combined merging/restructuring operation that enhances the action of the method in the tasks of approximating error minimizing.

The main advantage of SI and more powerful ASI method that it effectively copes with far from the optimal pixel clustering or image segmentation, providing more impressive improvement in the visual perception and also by the approximating error for the rougher initial image approximations [14].

## 6. EXPERIMENTAL RESULTS

Fig. 2 illustrates the effect of the combined merging/restructuring operation on the example of the standard color «Lena» image of 512x512 pixels, shown in the upper left corner. The *trivial* approximation of original image, which consists of the same pixels of a single color, constituting a single cluster is shown next to the right. Under the original image in the left column the approximations that are obtained by ordinary merging of adjacent segments in version [3,4,11] of Mumford-Shah segmentation model are placed. These contain from 2 to 5 segments of different colors. In the right column the appropriate image approximations with 1-5 colors, which are generated through combined merging/restructuring are placed. They are obtained by the same merging of adjacent segments, followed by the immediate restructuring of the sub-images, occupying these segments, to satisfy the convexity condition (5).

Fig. 2 demonstrates an example of conversion of segmentation to clustering. The effect of hierarchical segmentation improvement seems obvious. It is expressed both in visual perception, and also in decrease of the standard deviations, written out under the approximations.

It is important that, in contrast to the conventional hierarchical segmentation (left column in Fig. 2), the similar meaningful objects (eyes, pupils, etc.) simultaneously appears in the approximations for hierarchical clustering (right column in Fig.2). This effect is not accidental and can be used in a processing of stereo-pairs to detect previously unspecified objects for subsequent matching of their feature points and calculating the distances [15].

It is also important, that quite similar clustering results to those in Fig.2 for color image "Lena" of 512x512 pixels, are reproduced for the same image, but in the gray scale and of reduced size of 256x256 pixels. So, we assume that the formal objects i.e. resultant sub-images, stably detected by computer as the parts of visually observable objects, can be treated as workpieces of meaningful objects, at least in a number of practical tasks. ASI-method speeds up the calculations, provides tuning and promotes the object detection.

A number of obtained dichotomous sequences of image approximations is described in Fig. 3 that illustrates the standard deviation $\sigma$ of approximations from the image, depending on the number of clusters $g$ shown in the range from one to thousand.

The uppermost curve describes the sequence of approximations obtained by conventional segment merging (left column in Fig.2). Lower intertwined curves describe the optimized approximations generated by combined merging/restructuring, used instead of conventional merging. These describe the time-consuming approximations, obtained by Ward's method and by simple merging followed by immediate restructuring (right column in Fig.2), as well as the approximations, obtained using ASI-method, which was executed for different sub-image numbers specified in the range from 100 to 1000.

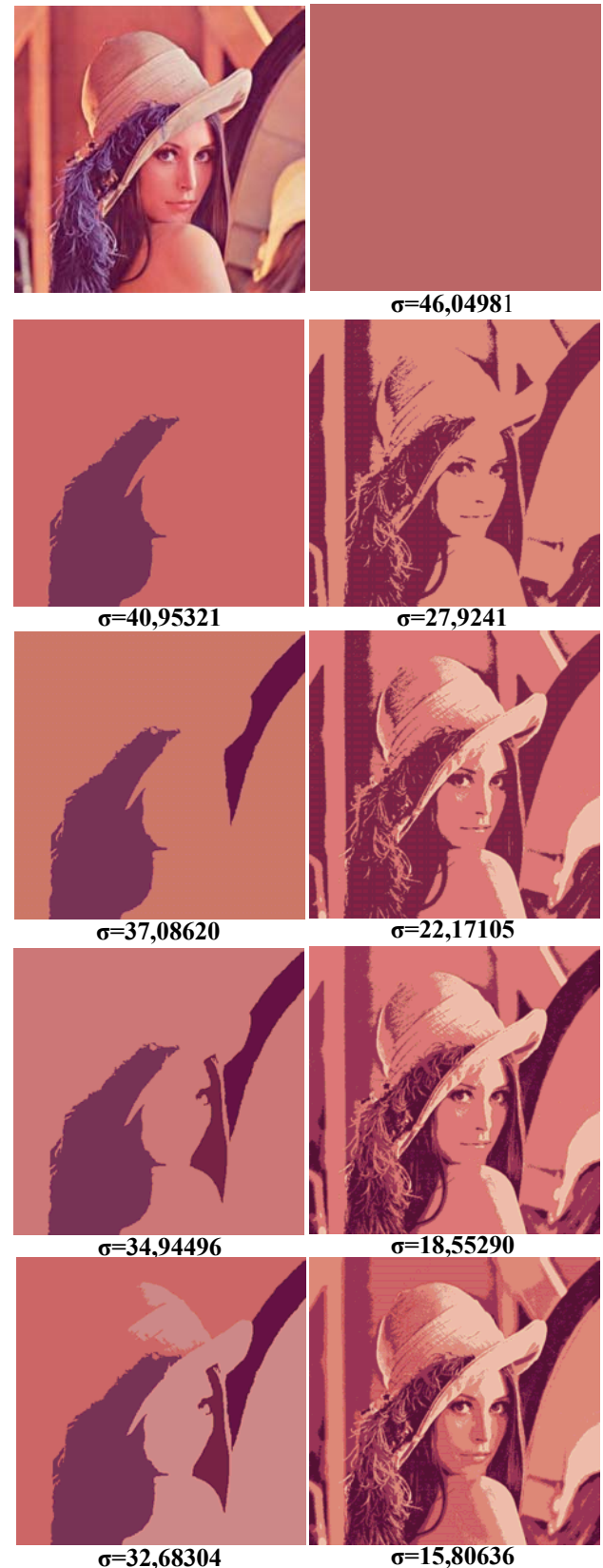

σ=46,04981

σ=40,95321    σ=27,9241

σ=37,08620    σ=22,17105

σ=34,94496    σ=18,55290

σ=32,68304    σ=15,80636

**Fig.2 − Segmentation-to-clustering conversion**

As can be seen in Fig. 3, the optimized approximations differ from the conventional much more than among themselves. As for visual perception all look natural and it is difficult to prefer one another. Apparently, it is useful to use several optimized dichotomous sequences of image approximations.

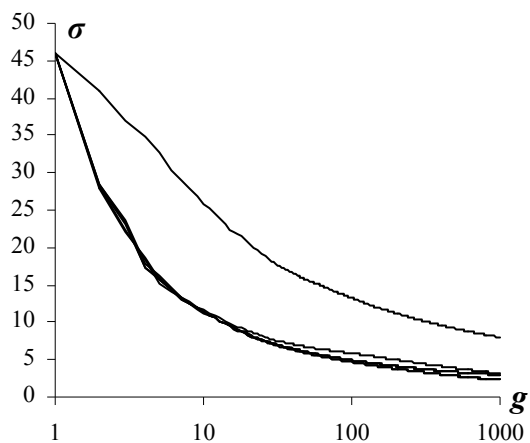

**Fig.3 – Standard deviation $\sigma$ depending on cluster number $g$ (in logarithmic scale)**

## 7. CONCLUSION

Thus, the paper presents the current development of the model of quasioptimal image approximations, intended for automation of low-level imaging using the classical cluster analysis. The task is to accurately simulate the image by approximations like that described in Fig.1. To achieve this, the combined merging/restructuring operation for hierarchically structured image formation and ASI-method for pixel clustering improvement are proposed in this paper.

The fact that improved hierarchical clustering or segmentation is reduced to simply replacing of basic merge operation by advanced combined operation is quite attractive for the unhindered implementation as in ASI-method. But for perfect minimizing of the approximating error, keeping the cluster numbers, ASI-method itself, especially in the case of a small cluster numbers, must be supplemented by proper version of so called K-meanless method [6,14], which, in fact, is the advanced version of K-means method [1].

The problem of excessive memory usage when operating with millions of image approximations is completely overcame, owing to data structure of Sleator-Tarjan dynamic trees [3,7,8]. However, when implementing and even a pilot study it is impossible to ignore the routine time-optimizing multi-iterative hierarchy generation, restructuring and optimization. But it's a one-time job. Therefore, it is likely that the problem of creating of an auxiliary tool for pixel clustering and image segmentation will be solved in the coming years. Then the development of the software will be engaged only in specific groups of developers, and the other specialists will be able to use ready-made programs.